\pdfoutput=1

\documentclass[11pt]{article}

\usepackage[]{emnlp2021}

\usepackage{times}
\usepackage{latexsym}

\usepackage[T1]{fontenc}

\usepackage[utf8]{inputenc}

\usepackage{microtype}

\usepackage{xcolor}
\usepackage{multirow}
\usepackage{tabularx}
\usepackage{booktabs}
\usepackage{graphicx}
\usepackage{url}
\usepackage{xspace}

\newcommand{\minds}{{\textsc{MInDS-14}}\xspace}

\title{Multilingual and Cross-Lingual Intent Detection from Spoken Data}

\author{
 Daniela Gerz\thanks{{ } Both authors equally contributed to this work.},
 Pei-Hao Su,$^{*}$
 Razvan Kusztos,
 Avishek Mondal,
 Michał Lis,
 Eshan Singhal,\\
 {\bf
 Nikola Mrk{\v{s}}i\'c,
 Tsung-Hsien Wen,
 Ivan Vuli\'{c}
 } \vspace{2mm} \\
 PolyAI Limited \\
 London, United Kingdom \\
 \texttt{\{dan,eddy,ivan\}@polyai.com}
}

\begin{document}
\maketitle
\begin{abstract}
We present a systematic study on multilingual and cross-lingual intent detection from spoken data. The study leverages a new resource put forth in this work, termed \minds, a first training and evaluation resource for the intent detection task with spoken data. It covers 14 intents extracted from a commercial system in the e-banking domain, associated with spoken examples in 14 diverse language varieties. Our key results indicate that combining machine translation models with state-of-the-art multilingual sentence encoders (e.g., LaBSE) can yield strong intent detectors in the majority of target languages covered in \minds, and offer comparative analyses across different axes: e.g., zero-shot versus few-shot learning, translation direction, and impact of speech recognition. We see this work as an important step towards more inclusive development and evaluation of multilingual intent detectors from spoken data, in a much wider spectrum of languages compared to prior work. 

\end{abstract}

\section{Introduction and Motivation}
\label{s:intro}
A crucial functionality of Natural Language Understanding (NLU) components in task-oriented dialogue systems is \textit{intent detection} \cite{Young:2002,Tur:2010slt,Coucke:18}. In order to understand the user's current goal, the system must classify their utterance into several predefined classes termed \textit{intents}. For instance, in the banking domain utterances referring to \textit{cash withdrawal}, \textit{currency exchange rates} or \textit{using credit cards abroad}  should be classified to the respective intent classes \cite{Casanueva:2020ws}. An error in intent detection is typically the first point of failure for any task-oriented dialogue system.

Scaling dialogue systems in general and intent detectors in particular to support a multitude of new dialogue tasks and domains is a challenging, time-consuming and resource-intensive process \cite{Wen:17,rastogi2019towards}. This problem is further exacerbated in \textit{multilingual setups}: it is extremely expensive to annotate sufficient task data in each of more than 7,000 languages \cite{Bellomaria:2019clic,Xu:2020emnlp}.\footnote{Further, reaching beyond the English language, other languages often exhibit different typological (e.g., morphosyntactic) and lexical properties, potentially requiring additional language-specific adaptations of English-trained models \cite{Ponti:2019cl,Hedderich:2021naacl}.}

As a consequence, the current work on intent detection has been largely constrained only to English, and standard intent detection benchmarks also exist only in English \cite[\textit{inter alia}]{Hemphill:1990,Larson:2019emnlp,Liu:2019iwsds,Casanueva:2020ws,Larson:2020emnlp}. The need to widen the reach of dialogue technology to other languages has been recognised only recently, and thus multilingual intent detection datasets are still few and far between: \newcite{Schuster:2019naacl} provide NLU data in three languages (English, Spanish, Thai), while a more recent MultiATIS++ dataset \cite{Xu:2020emnlp} manually translates the well-known ATIS dataset \cite{Hemphill:1990} from English to 8 target languages. Despite these first efforts, there are still prominent gaps remaining: \textbf{1)} a large number of (even major) languages is still uncovered, \textbf{2)} there are no multilingual data for specialized and well-defined domains such as e-banking, and \textbf{3)} most importantly, all intent detection datasets to date are text-based. In other words, current work completely ignores the fact that many conversational systems are inherently \textit{voice-based}, and that telephony quality and errors in automatic speech recognition (ASR) even prior to intent detection may have fundamental impact on the final intent detection performance. Consequently, the impact of ASR on multilingual intent detection has not been studied before. 

\vspace{1.4mm}
\noindent \textbf{Contributions.} Inspired by the current gaps, \textbf{1)} we present the \textsc{MInDS-14} dataset (\textbf{M}ultilingual \textbf{In}tent \textbf{D}etection from \textbf{S}peech), a first multilingual evaluation resource for intent detection from spoken data. The dataset originates from the use of a commercial voice assistant and real-life industry needs: it covers 14 intents in the banking domain in 14 different language varieties, making it the most comprehensive multilingual intent detection dataset to date. \textbf{2)} We present a systematic evaluation and comparison of current state-of-the-art multilingual and cross-lingual intent detection models, which rely on machine translation and current cutting-edge multilingual sentence encoders such as multilingual USE \cite{Chidambaram:2019repl} and LaBSE \cite{Feng:2020labse}. \textbf{3)} We provide a series of additional analyses and experiments to further profile the potential and current gaps of intent detection in multilingual voice-base contexts, including few-shot versus zero-shot transfer, target-only versus multilingual training, and aggregations of n-best ASR hypotheses. 

Our results demonstrate that strong intent detection results can be achieved for all languages represented in \minds, but we also indicate the crucial importance of in-domain model fine-tuning and few-shot learning, reporting strong gains over zero-shot transfer models. In hope to motivate and inspire further work on multilingual \textit{and} voice-based intent detection, we release the \minds dataset, inclucing the original speech data and ASR data, online at: {\small \url{s3://poly-public-data/MInDS-14/MInDS-14.zip}}.

\section{\minds: Dataset Collection}
\label{ss:collection}

\noindent \textbf{Final Dataset and Languages Covered.}
The final \minds dataset covers 14 intents in the banking domain with accompanying spoken and ``ASR-ed'' utterances. The intents were sampled from a set of 90+ fine-grained intents used by a commercial banking voice assistant, so that all intents have a clear and non-overlapping semantics, and are easy to understand by non-experts, i.e., crowdsourcers. Around 50 examples for all 14 intents are collected in 14 different language varieties, with the exact numbers available in the appendix. The language set includes \textbf{a)} three varieties of English: British (\textsc{en-gb}), US (\textsc{en-us}), and Australian (\textsc{en-au}); \textbf{b)} Germanic and Romance Western European languages: French (\textsc{fr}), Italian (\textsc{it}), Spanish (\textsc{es}), Portuguese (\textsc{pt}), German (\textsc{de}), and Dutch (\textsc{nl}); \textbf{c)} Slavic: Russian (\textsc{ru}), Polish (\textsc{pl}), Czech (\textsc{cs}); and \textbf{d)} Asian languages: Korean (\textsc{ko}), Chinese (\textsc{zh}).\footnote{The choice of languages has been driven by typological diversity \cite{Ponti:2019cl}, but also by the number of native speakers and the number of available participants on the used crowdsourcing websites.}

\vspace{1.4mm}
\noindent \textbf{Spoken Data Collection.} 
The spoken data has been collected via crowdsourcing, relying on the Prolific platform.\footnote{\url{https://www.prolific.co/}} We have experimented with two different data collection protocols, which eventually yield very similar data quality. With both protocols, human subjects are first provided with the particular intent class, a description of the intent, and three examples for the intent class. The task is then to provide new spoken utterances associated with the intent class. 

As the first collection protocol, we implement a full-fledged phone-based voice assistant that participants could call and talk to. This approach makes the data collection setup as realistic as possible: it is affected by the (phone) audio quality and directly captures the way people would speak on the phone. \textsc{it} data and parts of \textsc{de}, \textsc{pt}, \textsc{pl}, and \textsc{en-au} data have been collected via this approach. The second, simpler study design instead relies on an online recording software. We use Phonic\footnote{\url{https://www.phonic.ai/}} to collect the recordings, where data collection for each intent class is set up as a dedicated task on Prolific. We collect all the other data items via this approach.

In order to ensure native pronunciation data quality, the pool of participants has been restricted to native speakers from the relevant regions. After the initial collection step, the data was inspected and cleaned manually to remove empty, nonsensical, and extremely long utterances.

\vspace{1.4mm}
\noindent \textbf{Auxiliary English Data.} Along the main multilingual dataset, we also release an auxiliary English intent detection dataset (termed \textsc{aux-en} henceforth) in the same banking domain. It comprises a total of 660 English utterances, extracted from a commercial voice assistant, and annotated with the same 14 intent classes. The dataset allows us to run cross-lingual transfer and training data augmentation experiments and analyses later in \S\ref{s:results}. 


\section{Multilingual and Cross-Lingual Intent Detection: Methodology}
\label{ss:mid}
A standard transfer learning paradigm \cite{Ruder:2019transfer} fine-tunes a pretrained language model such as BERT \cite{Devlin:2018arxiv} or RoBERTa \cite{Liu:2019arxiv} on the annotated task data. For the intent classification task in particular, \newcite{Casanueva:2020ws} have recently shown that full fine-tuning of the large pretrained model is not needed at all. In contrast, they propose a more efficient \textit{feature-based} approach to intent detection. Here, fixed universal sentence encoders such as USE \cite{Cer:2018arxiv,Chidambaram:2019repl} or ConveRT 
\cite{Henderson:2020convert} are used ``off-the-shelf'' to encode utterances, and a standard multi-layer perceptron (MLP) classifier is then learnt on top of the sentence encodings. \newcite{Casanueva:2020ws} demonstrate that the feature-based approach to intent classification yields performance on-par with the full-model fine-tuning, while offering improved training efficiency. Therefore, due to the large number of executed experiments and comparisons in this work, we opt for this efficient approach to intent detection, and outline it in what follows.

\vspace{1.4mm}
\noindent \textbf{Pretrained Sentence Encoders.} 
We evaluate two widely used state-of-the-art multilingual sentence encoders. Here, we provide only brief descriptions of each encoder; for more details regarding each model, we refer the reader to the original work.

\vspace{1.2mm}
\noindent \textbf{mUSE.} Multilingual Universal Sentence Encoder (mUSE) \cite{yang2020mUSE} is a multilingual version of the Universal Sentence Encoder (USE) model for English \cite{Cer:2018arxiv}.  It relies on a standard dual-encoder neural framework \cite{Henderson:2019acl,Reimers:2019emnlp,Humeau:2019arxiv}. It features 16 languages, and it learns a shared cross-lingual semantic space via translation-bridging tasks \cite{Chidambaram:2019repl}. Following \newcite{Litschko:2021ecir}, we opt for this mUSE variant: 3-layer Transformer encoder with average-pooling.

\vspace{1.2mm}
\noindent \textbf{LaBSE.} Language-agnostic BERT Sentence Embedding (LaBSE) \cite{Feng:2020labse} adapts pretrained multilingual BERT (mBERT) \cite{Devlin:2018arxiv} using a dual-encoder framework \cite{yang2019ijcai} with larger embedding capacity (i.e., it provides a shared multilingual vocabulary spanning 500k subwords). In addition to the multi-task training objective of mUSE, LaBSE uses standard self-supervised objectives used in pretraining of mBERT and XLM: masked and translation language modeling \cite{Conneau:2019nips}. LaBSE is the current state-of-the-art multilingual sentence encoder, and supports 109 languages.

\vspace{1.2mm}
We keep all pretrained sentence encoders fixed during intent detection training. Formally, we pass a user utterance, that is, a sequence of input tokens $x = (x_0,x_1,...,x_T)$ through an encoder model $\theta_{enc}$, where $\theta_{enc}$ is one of the models described above, producing the sequence encoding $s_x=\theta_{enc}(x)=\theta_{enc}((x_0,x_1,...,x_T))$.
%

\vspace{1.4mm}
\noindent \textbf{Intent Classification Model.} For intent detection, we pass the sentence encoding $s_x$ through a 2-layer MLP. We first apply dropout \cite{Srivastava:2014} on the sentence vector, followed by one layer with ReLU as nonlinear activation \cite{Nair:2010icml}, yielding the hidden representation $h = ReLU(W_1 s_{dp}+b_1)$, where $W_1$ is a trainable weight matrix, $s_{dp}$ is the encoding after applying dropout, and $b_1$ denotes bias parameters.

 


We then detect the intent using a sigmoid ($\sigma$) activation and softmax classification as follows:
\begin{equation}
   p_{intent} = softmax(\sigma (W_2h+b_2)),
\end{equation}
\noindent where $W_2$ is another trainable weight matrix, and $b_2$ are bias parameters.

\section{Experimental Setup}
\label{s:exp}
\noindent \textbf{Speech Transcription.} For all recordings in all languages, we run the Google ASR model\footnote{\url{cloud.google.com/speech-to-text}} for the respective language variant to obtain $n$-best written transcriptions (i.e., ASR hypotheses). Unless noted otherwise, we assume working only with the top (i.e., 1-best) written transcription.

\vspace{1.4mm}
\noindent \textbf{Monolingual versus Multilingual Training.}
We then train and run the intent classification models based on sentence encoders from \S\ref{ss:mid} in the following setups. First, in the \textbf{translate-to-EN} variant, for all ``non-English'' languages, we translate the transcriptions into English via Google Translate. This effectively enables us to train and evaluate monolingual models directly in English, using the pretrained multilingual (or English-only) sentence encoders. The second approach works directly in the native language of the transcriptions, and we discern between two variants: \textbf{a) target-only} uses only the data available in the current language to train the intent classifier, while in the \textbf{b) multilingual} setup we leverage the multilinguality of mUSE and LaBSE and train on the transcribed data of all languages, while we evaluate on the test data of each individual language.

\begin{figure}[!t]
\centering
\includegraphics[width=0.87\linewidth]{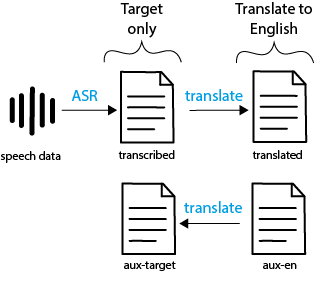}
\vspace{-2mm}
\caption{Illustration of different training and evaluation data and scenarios.}
\label{figure:setup}
\vspace{-2mm}
\end{figure}



\begin{table*}[!t]
\begin{center}
{\footnotesize
\begin{tabularx}{\linewidth}{l XXXX XXXX X}
    \toprule
    {} &  \multicolumn{4}{c}{\textbf{translate-to-EN}} & \multicolumn{4}{c}{\textbf{target-only}} & \multicolumn{1}{c}{\textbf{multilingual}} \\
    \cmidrule(lr){2-5} \cmidrule(lr){6-9} \cmidrule(lr){10-10}
    {} & {\it aux-o}  & {\it no-aux} &  {\it standard} & {\it standard} &  {\it aux-o} & {\it aux-o} &  {\it standard} &  {\it standard} & {\it standard} \\
    \textbf{Language} & LaBSE & LaBSE &  mUSE & LaBSE &  mUSE & LaBSE &  mUSE &  LaBSE & LaBSE \\
    \cmidrule(lr){2-5} \cmidrule(lr){6-9} \cmidrule(lr){10-10}
    \textsc{cs} & {68.0} & {95.9} & {90.1} & {95.7} & {(34.4)} & {63.6} & {(69.6)} & {94.9} & {91.7} \\
    \textsc{de} & {69.6} & {95.6} & {91.0} & {95.0} & {51.1} & {53.6} & {89.2} & {93.2} & {94.2} \\
    \textsc{en-au} & {77.1} & {95.9} & {93.4} & {94.4} & {61.2} & {74.8} & {96.1} & {93.7} & {94.5} \\
    \textsc{en-gb} & {73.6} & {96.4} & {91.6} & {94.7} & {55.8} & {75.3} & {94.1} & {96.5} & {94.4} \\
    \textsc{en-us} & {76.7} & {95.1} & {97.1} & {95.7} & {57.8} & {80.1} & {94.6} & {95.1} & {95.3} \\
    \textsc{es} & {68.7} & {95.8} & {95.8} & {92.2} & {49.6} & {62.7} & {87.5} & {91.9} & {91.5} \\
    \textsc{fr} & {75.3} & {97.1} & {93.1} & {94.3} & {62.4} & {62.5} & {92.6} & {93.1} & {92.6} \\
    \textsc{it} & {71.4} & {97.4} & {93.4} & {95.8} & {56.3} & {65.6} & {85.8} & {96.2} & {92.3} \\
    \textsc{ko} & {73.5} & {94.0} & {86.3} & {91.1} & {53.0} & {65.6} & {84.6} & {91.4} & {90.5}\\
    \textsc{nl} & {67.7} & {95.8} & {94.0} & {92.4} & {53.8} & {58.1} & {85.6} & {91.0} & {96.5} \\    
    \textsc{pl} & {76.6} & {93.7} & {81.9} & {94.9} & {51.3} & {45.7} & {80.3} & {89.2} & {93.8}\\ 
    \textsc{pt} & {69.7} & {97.5} & {90.5} & {94.4} & {55.3} & {53.1} & {96.8} & {95.3} & {92.7}\\ 
    \textsc{ru} & {68.0} & {95.7} & {93.7} & {95.1} & {43.1} & {68.9} & {88.4} & {93.6} & {95.2}\\
    \textsc{zh} & {72.5} & {96.1} & {86.6} & {95.6} & {53.2} & {62.7} & {81.6} & {93.0} & {90.8}\\
    \cmidrule(lr){2-5} \cmidrule(lr){6-9} \cmidrule(lr){10-10}
    {\bf Average} & {72.0} & {95.9} & {91.3} & {94.4} & {52.7} & {63.7} & {87.6} & {93.4} & {93.3}\\
    \bottomrule
\end{tabularx}
}%
\end{center}
\vspace{-1mm}
\caption{Main results on the \minds benchmark, with different training and evaluation setups (see \S\ref{s:exp}). Accuracy scores are reported. \textit{aux-o} refers to the \textit{aux-only} training setup; the scores for Czech in the target-only setup with mUSE are in parentheses because mUSE has not been trained with any Czech data.}
\label{tab:main_results}
\vspace{-1mm}
\end{table*}

\vspace{1.4mm}
\noindent \textbf{Training and Evaluation Data and Setups.} 
We can also translate our auxiliary English dataset (\textsc{aux-en}, see \S\ref{ss:collection}) to all other languages via Google Translate, yielding auxiliary \textsc{aux-target} data. We then discern between the following training data setups. In \textbf{a) aux-only} we use only the \textsc{aux-target} (or \textsc{aux-en}) data to train the intent classifier. This setting allows us to estimate the intent detection performance before any additional in-language data collection. In \textbf{b)} the \textbf{standard} setup, we do 3-fold cross-validation, where we randomly split the transcribed data (translate-to-EN, target-only, or multilingual) into 60\% training data and 40\% test data, and always add the auxiliary data to the training subset.\footnote{In the \textit{aux-only} variant we still sample 40\% of the entire dataset for testing in order to work with the test sets of the same size as in the \textit{standard} variant. For \textit{multilingual} training, in order to maintain the same multilingual training set for all test languages, we also sample 60\% of all transcribed data in all languages, and use that plus all \textsc{aux-target} data for training, and the remaining 40\% in each language for testing.} For a subset of experimental runs, we also evaluate the \textbf{c) no-aux} setup, where we train only on the 60\% of the in-domain data, without any auxiliary data. An illustration of these different setups is provided in Figure~\ref{figure:setup}.


\vspace{1.4mm}
\noindent \textbf{Hyperparameters.} 
The intent classifiers using mUSE and LaBSE as sentence encoders are configured as follows. We train with Adam \cite{adam:15} relying on the learning rate of 0.001, in batches of size 32, for 10,000 steps. The dropout rate \cite{Srivastava:2014} is set to 0.3. We report accuracy as the main evaluation measure for all experimental runs, always averaged over 3 independent runs.


\section{Results and Discussion}
\label{s:results}
The main results are summarised in Table~\ref{tab:main_results}, and they offer several axes of comparison, discussed in what follows. First, the results confirm that LaBSE is a stronger multilingual encoder across the board, extending its superiority over mUSE from cross-lingual sentence matching tasks \cite{Feng:2020labse} also to multilingual intent detection. 

More importantly, the results indicate very high absolute accuracy scores for all target languages, confirming the validity of MT-based approaches to multilingual intent detection, at least for major languages with developed MT. For instance, the results for all languages are $>95\%$ (except for Korean and Polish) with LaBSE as the supporting sentence encoder in the \textit{no-aux} translate-to-EN setup. In other words, we empirically demonstrate the viability of the simple ``ASR-then-translate'' approach when dealing with voice-based input, at least for the current group of languages, which are all considered reasonably high-resource in NLP terms. Our findings suggest that even this simple, easy-to-build, and efficient sentence encoder-based approach may offer competitive intent detectors from spoken data in different languages. Future work will investigate the extent of performance drops once the focus is shifted to lower-resource languages where reasonably performing ASR and MT models cannot be guaranteed \cite{Conneau:2020asr,Pratap:2020:50}, as well as to finer-grained intent classes and other domains. 

A comparison of different training and data setups reveals that even small in-domain training data (without any external data augmentation, the \textit{no-aux} setup) are sufficient to learn strong intent detectors. In fact, the best overall results are achieved with the \textit{no-aux} translate-to-EN setup with LaBSE, having a slight edge over the setup which additionally employs the \textsc{aux-en} data. The \textit{aux-only} setup where we learn intent detectors on a very related domain fall substantially behind in-domain trained models, validating the crucial importance of few-shot learning (i.e., using even the small portions of fully in-domain training data to boost performance). We also note that slightly better results on average are achieved in the ``translate-to-EN'' scenario than in the ``target-only'' and 'multilingual'' scenarios. However, the differences when using LaBSE are slight, and there are some languages with higher scores achieved in the other two scenarios. Interestingly, we do not observe any boosts on average with data augmentation in the multilingual setup. This warrants further investigation in future work.

\subsection{Further Discussion}
\label{ss:further}


\noindent \textbf{Performance Variance.} We remind the reader that the reported scores are the average across 3 runs, and further note that performance may vary between different runs with the same hyperparameters, which is a known problem when fine-tuning large pretrained models with small amounts of task data \cite{Phang2018, ruder2021lmfine-tuning}. For instance, variance for the target-only \textit{standard} setup with LaBSE is at 3.8 accuracy points on average. This is also the reason why we have mostly focused on high-level trends in the discussion of the results, and why we always average the scores over several independent runs with the same hyperparameters \cite{Dodge:2019emnlp}. 

\begin{figure}[!t]
\includegraphics[width=0.95\linewidth]{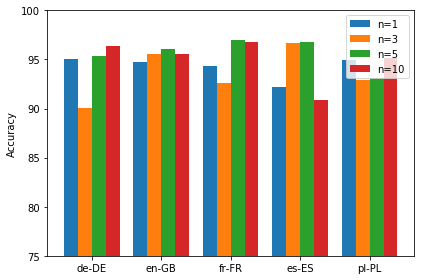}
\caption{Results with added training data from the ASR $n$-best list. Target-only \textit{standard}; LaBSE.}
\label{figure:asr}
\end{figure}

\vspace{1.4mm}
\noindent \textbf{Impact of ASR.} For a subset of languages, we also evaluate whether including additional ASR hypotheses might make intent detectors more robust: adding more transcriptions from the $n$-best list may be seen as a form of data augmentation. The results with adding $n=1,3,5,10$ examples are provided in Figure~\ref{figure:asr}.\footnote{As before, for test examples we always take the best ($n=1$) transcription.} The scores suggest that relying on more transcriptions ($n=5$ and $n=10$) does yield slight gains on average, but the trend is not present in all the test languages (cf., Spanish). This might stem from the fact that the transcriptions are highly similar, and there is limited additional information available down the $n$-best ASR list. 


\begin{figure}[!t]
\includegraphics[width=0.95\linewidth]{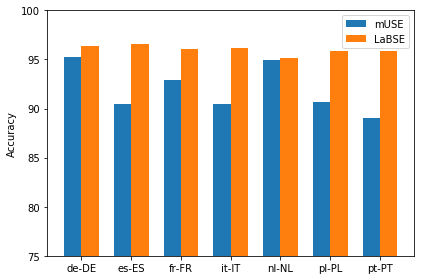}
\caption{Results when training with translations obtained from two translation services: Google Translate and DeepL. Translate-to-EN \textit{standard} setup.}
\label{figure:translate}
\end{figure}


\vspace{1.4mm}
\noindent \textbf{Impact of Additional Translations.}
Another approach to improving robustness of the intent detectors is to generate more than one (machine) translation per each transcription. We achieve that by passing each transcription from the target language not only through Google Translate, but also through another translation service: DeepL,\footnote{\url{https://www.deepl.com/en/translator}} yielding 2 translations per each transcription.\footnote{This can also be seen as a form of data augmentation.} The results in the \textit{standard} translate-to-EN setup with LaBSE and mUSE are provided in Figure~\ref{figure:translate}. They indicate that this ``augmentation via translation'' step indeed yields slightly improved intent detectors: we hit 1-2\% performance gains with both sentence encoders (cf., Figure~\ref{figure:translate} and Table~\ref{tab:main_results}) compared to using only 1 one translation per transcription.



\section{Conclusion and Future Work}
\label{s:conclusion}
We have presented a first study focused on multilingual and cross-lingual intent detection from spoken data. To this end, we have presented \minds, a first training and evaluation resource for the task with spoken data, covering 14 intents extracted from a commercial system in the e-banking domain, with spoken examples available in 14 language varieties. Our key results have revealed that it is possible to build accurate intent detectors in all target languages relying on a simple yet efficient paradigm based on current state-of-the-art multilingual sentence encoders such as LaBSE and machine translation. In future work we plan to expand the \minds dataset and put more focus on similar evaluations for truly low-resource languages, where reliable ASR, MT, and even sentence encoders cannot be guaranteed. We also plan to experiment with finer-grained intent classes and build similar resources in other domains.

In the long run, we hope that our initiative will foster future developments and evaluation of multilingual intent detectors from spoken data, as one of the first steps towards truly multilingual voice-based conversational AI.

\section*{Acknowledgements}
We are grateful to our colleagues at PolyAI for many fruitful discussions and suggestions.

\bibliography{custom}
\bibliographystyle{acl_natbib}

\newpage
\appendix
\section{Appendix}

\subsection{List of Intents}

\textsc{Business loan} \\
\textsc{Freeze} \\
\textsc{Abroad} \\
\textsc{App error} \\
\textsc{Direct Debit} \\
\textsc{Card issues} \\
\textsc{Joint Account} \\
\textsc{Balance} \\
\textsc{High value payment}\\
\textsc{Atm limit} \\
\textsc{Address} \\
\textsc{Pay bill} \\
\textsc{Cash deposit} \\
\textsc{Latest transactions} \\

\subsection{Stats}
\label{sec:appendix}
\begin{table}[h!]
\begin{center}
\begin{tabularx}{\linewidth}{l | X }
    \hline
    Language & Number of Examples \\
    \hline
    \textsc{cs} & {574} \\
    \textsc{de} & {611} \\
    \textsc{en-AU} & {654} \\
    \textsc{en-GB} & {592} \\
    \textsc{en-US} & {563} \\
    \textsc{es} & {486} \\
    \textsc{fr} & {539} \\
    \textsc{it} & {696} \\
    \textsc{ko} & {592} \\
    \textsc{nl} & {654} \\
    \textsc{pl} & {562} \\ 
    \textsc{pt} & {604} \\ 
    \textsc{ru} & {539} \\
    \textsc{zh} & {502} \\
    \hline
\end{tabularx}
\end{center}
\caption{Number of examples per language.}
\label{tab:stats}
\end{table}

\end{document}